\documentclass{article}
\usepackage{ijcai17}

\usepackage{times}
\usepackage{helvet}
\usepackage{courier}
\usepackage{multirow}
\usepackage{latexsym}
\usepackage{mathrsfs}
\usepackage{graphicx}
\usepackage{subfig}
\usepackage{amsfonts}
\usepackage{amssymb}
\usepackage{amsmath}
\usepackage{bm}
\usepackage{url}
\usepackage{tikz-dependency}
\usepackage{pgfplots}
\usepackage{xcolor}
\usepackage{array}
\usepackage{enumitem}
\usepackage{booktabs}
\newcommand\newcite[1]{\citeauthor{#1} [\citeyear{#1}]}
\usepackage{booktabs}

\allowdisplaybreaks

\newenvironment{itemize*}%
 {\begin{itemize}%
  \setlength{\itemsep}{0pt}%
  \setlength{\parskip}{0pt}}%
 {\end{itemize}}
 \newenvironment{enumerate*}%
 {\begin{enumerate}%
  \setlength{\itemsep}{0pt}%
  \setlength{\parskip}{0pt}}%
 {\end{enumerate}}

\DeclareMathOperator*{\softmax}{softmax}
\DeclareMathOperator*{\Basic}{DC-TreeNN}
\def\W{\mathbf{W}}

\def\bb{\mathbf{b}}

\def\h{\mathbf{h}}

\def\cc{\mathbf{c}}

\def\cc{\mathbf{c}}

\def\D{\mathbf{D}}
\def\z{\mathbf{z}}
\def\x{\mathbf{x}}

\title{Dynamic Compositional Neural Networks over Tree Structure}
\author{Pengfei Liu \quad Xipeng Qiu\thanks{Corresponding author.} \quad Xuanjing Huang\\
Shanghai Key Laboratory of Intelligent Information Processing, Fudan University\\
School of Computer Science, Fudan University\\
825 Zhangheng Road, Shanghai, China\\
\{pfliu14,xpqiu,xjhuang\}@fudan.edu.cn}

\begin{document}
\maketitle
\begin{abstract}
Tree-structured neural networks have proven to be effective in learning semantic representations by exploiting
syntactic information. In spite of their success, most existing models suffer from the underfitting problem: they recursively use the same shared compositional function throughout the whole compositional process and lack expressive power due to inability to capture the richness of compositionality.
In this paper, we address this issue by introducing the dynamic compositional neural networks over tree structure (DC-TreeNN), in which the compositional function is dynamically generated by a meta network.
The role of meta-network is to capture the metaknowledge across the different compositional rules and formulate them. Experimental results on two typical tasks show the effectiveness of the proposed models.
\end{abstract}

\section{Introduction}

Learning the distributed representation for long spans of text from its constituents has been a key step for various natural language processing (NLP) tasks, such as text classification \cite{zhao2015self,liu2015multitimescale}, semantic matching \cite{liufusion,liucouple2016}, and machine translation \cite{cho2014learning}.
%Learning the distributed representation for long spans of text from its constituents has been a key problem in natural language processing (NLP).
Existing deep learning approaches take a compositional function with different forms
to compose word vectors recursively until obtaining a sentential representation.
Typically, these compositional functions involve recurrent neural networks \cite{hochreiter1997long,sutskever2014sequence}, convolutional neural networks \cite{collobert2011natural,kalchbrenner2014convolutional}, and tree-structured neural networks \cite{tai2015improved,zhu2015long}.

Among these methods,  tree-structured neural networks (Tree-NNs) show theirs superior performance in many NLP tasks \cite{socher2012semantic,irsoy2014deep}.
Following the syntactic tree structure, Tree-NNs assign a fixed-length vector to each word at the leaves of the tree, and combine word and phrase pairs recursively to create intermediate node vectors, eventually obtaining one final vector to represent the whole sentence.

However, these models have a major limitation in their inability to fully capture the richness of compositionality \cite{socher2013parsing}. The same parameters are used for all kinds of semantic compositions, even though the compositions have different characteristics in nature. For example, the composition of the adjective and the noun differs significantly  from the composition of the verb and the noun.
Moreover, many semantic phenomena, such as semantic idiomaticity or transparency, call for more powerful compositional mechanisms \cite{pylkkanen2006syntax}. Therefore, Tree-NNs suffer from the underfitting problem.

To alleviate this problem, some researchers propose to use multiple compositional functions, which are arranged beforehand according to some partition criterion \cite{socher2012semantic,socher2013parsing,dong2014adaptive}. Intuitively, using different parameters for different types of compositions has the potential to greatly reduce underfitting.
\newcite{socher2013parsing} defined different compositional functions in terms of syntactic categories, and a suitable compositional function is selected based on the syntactic categories.
\newcite{dong2014adaptive} introduced multiple compositional functions and during compositional phase, a proper one is selected based on the input information.
Although these models accomplished their mission to a certain extent, they still suffer from the following three challenges. First, the predefined compositional functions cannot cover all the compositional rules; Second, they require more learnable parameters, suffering from the problem of overfitting; Third, it is difficult to determine a universal criterion for semantic composition based solely on syntactic categories.

In this paper, we propose dynamic compositional neural networks over tree structure, in which
a \textit{meta network} is used to generate the context-specific parameters of a \textit{dynamic compositional network}. Specifically, we construct our models based on two kinds of tree-structured neural networks: recursive neural network (Tree-RecNN) \cite{socher2012semantic} and tree-structure long short-term memory neural network (Tree-LSTM) \cite{tai2015improved}. Our work is inspired by recent work on dynamic parameter prediction \cite{de2016dynamic,bertinetto2016learning,ha2016hypernetworks}. The meta network is used to extract the shared meta-knowledge across different compositional rules and to dynamically generate the context-specific compositional function. Thus, the compositional function of our models varies with positions, contexts and samples. The dynamic compositional network then applies those context-specific parameters to the current input information. Both meta and dynamic networks are differentiable such that the overall networks can be trained in an end-to-end fashion.
Additional, to reduce the complexity of the whole networks, we define the dynamic weight matrix in a manner simulating low-rank matrix decomposition.

We evaluate our models on two typical tasks: text classification and text semantic matching.
The results show that our models are more expressive due to their learning to learn nature, yet without increasing the number of model's parameters.
Moreover, we find certain composition operations can be learned implicitly by meta TreeNN, such as the composition of noun phrases and verb phrases.

The contributions of the paper can be summed up as follows.
\begin{enumerate}
    \item We provide a new perspective on how to compose the individual word of a sentence. Instead of directly using a learnable parameterized compositional function, we introduce a meta neural network, which can generate a compositional network to dynamically compose constituents over tree structure.
    \item Experimental results show that the proposed architecture is more expressive due to its learning to learn nature, yet without increasing the number of model's parameters.
    \item We present an elaborate qualitative analysis, giving an intuitive understanding on how our model works from semantic and syntactic perspectives.
\end{enumerate}

\section{Tree-Structured Neural Network}

In this section, we briefly describe the tree-structured neural networks.

The idea of tree-structured neural networks for natural language processing (NLP) is to train a deep learning model with a grammatical tree structure \cite{pollack1990recursive} that can be applied to phrases and sentences.
At every node in the tree, the contexts of the left and right children are combined by a compositional function. The parameters of the compositional function are shared across all nodes in the tree. The layer computed at the top node gives a representation for the whole sentence.

\subsection{Vanilla Recursive Neural Network}
The simplest member of tree-structured NN is the vanilla recursive neural network \cite{socher2013parsing}, in which
the representation of parent is calculated by weighted linear combination of the child vectors.

Formally, given a binary constituency tree $T$ induced by a sentence, each non-leaf node corresponds to a phrase. We refer to $\h_{j} \in \mathbb{R}^{d}$ as the hidden state of each node $j$, and let $\mathbf{h}_{j}^{l}$, $\mathbf{h}_{j}^{r}$ denote the left and right child representations respectively.
\begin{align}
    \h_{j} = \tanh
    \left(
	\mathbf{W}
	\begin{bmatrix}
		%\mathbf{x}_{j} \\
		\mathbf{h}_{j}^{l} \\
    \mathbf{h}_{j}^{r} \\
	\end{bmatrix} + \mathbf{b} \right),\label{eq:RecNN}
\end{align}
where $\mathbf{W} \in \mathbb{R}^{d \times 2d}$ is a learnable compositional matrix, $\mathbf{b}$ is the bias vector.

%Multiple recursive compositional functions have been explored, from linear transformation matrices to tensor products \cite{socher2012semantic,socher2013parsing}.
\begin{figure}[t]\centering
 \includegraphics[width=0.45\linewidth]{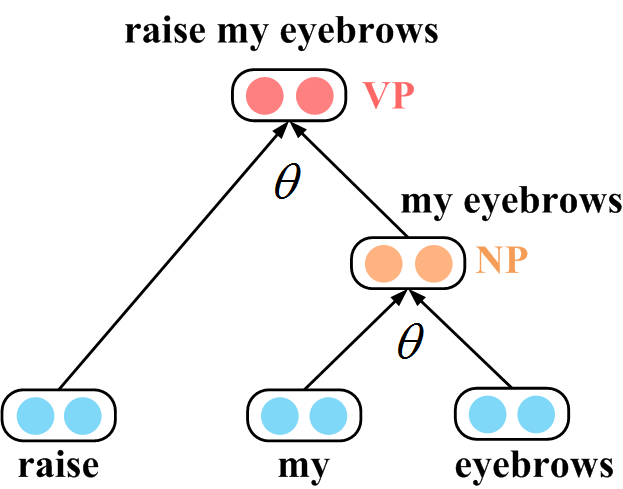}
 \caption{Illustration of Vanilla Tree Structure Network. NP and VP represent noun and verb phrases respectively. $\theta$ denotes the shared parameters of compositional function.
 }\label{fig:match}
\end{figure}

\subsection{Tree LSTM}
Tree LSTM \cite{tai2015improved} is a generalization of LSTMs to tree-structured network topologies.
In this model, the compositional function is an LSTM unit, and the hidden state $h_j$ of each node can be computed as follows:
we refer to $\h_{j}$ and $\cc_{j}$ as the hidden state and memory cell of each node $j$. The transition equations of node $j$ are as follows:

\begin{align}
	\begin{bmatrix}
		\mathbf{\tilde{c}}_{j} \\
		\mathbf{o}_{j} \\
		\mathbf{i}_{j} \\
		\mathbf{f}_{j}^{l} \\
    \mathbf{f}_{j}^{r}
	\end{bmatrix}
	&=
	\begin{bmatrix}
		\tanh \\
		\sigma \\
		\sigma \\
		\sigma \\
    \sigma
	\end{bmatrix}
    \left(
	\mathbf{W}
	\begin{bmatrix}
		\mathbf{x}_{j} \\
		\mathbf{h}_{j}^{l} \\
    \mathbf{h}_{j}^{r} \\
	\end{bmatrix} + \mathbf{b} \right),\label{eq:treelstm1}\\
    \mathbf{c}_{j} &=
		\mathbf{\tilde{c}}_{j} \odot \mathbf{i}_{j}
		+ \mathbf{c}_{j}^{l} \odot \mathbf{f}_{j}^{l}+ \mathbf{c}_{j}^{r} \odot \mathbf{f}_{j}^{r} , \\
	\mathbf{h}_{j} &= \mathbf{o}_{j} \odot \tanh\left( \mathbf{c}_{j} \right)\label{eq:treelstm3},
\end{align}
where $\mathbf{x}_j \in \mathbb{R}^{d}$ denotes the input vector and is non-zero if and only if it is a leaf node. The superscript $l$ and $r$ represent the left child and right child respectively.
$\sigma$ represents the logistic sigmoid function and $\odot$ denotes element-wise multiplication.
$\mathbf{W} \in \mathbb{R}^{5d \times 3d}$ and $\mathbf{b} \in \mathbb{R}^{5d}$ are learnable parameters.

\section{Dynamic Compositional Neural Network}

\begin{figure}[t]
\setlength{\belowcaptionskip}{-0.3cm} \centering
 \includegraphics[width=0.8\linewidth]{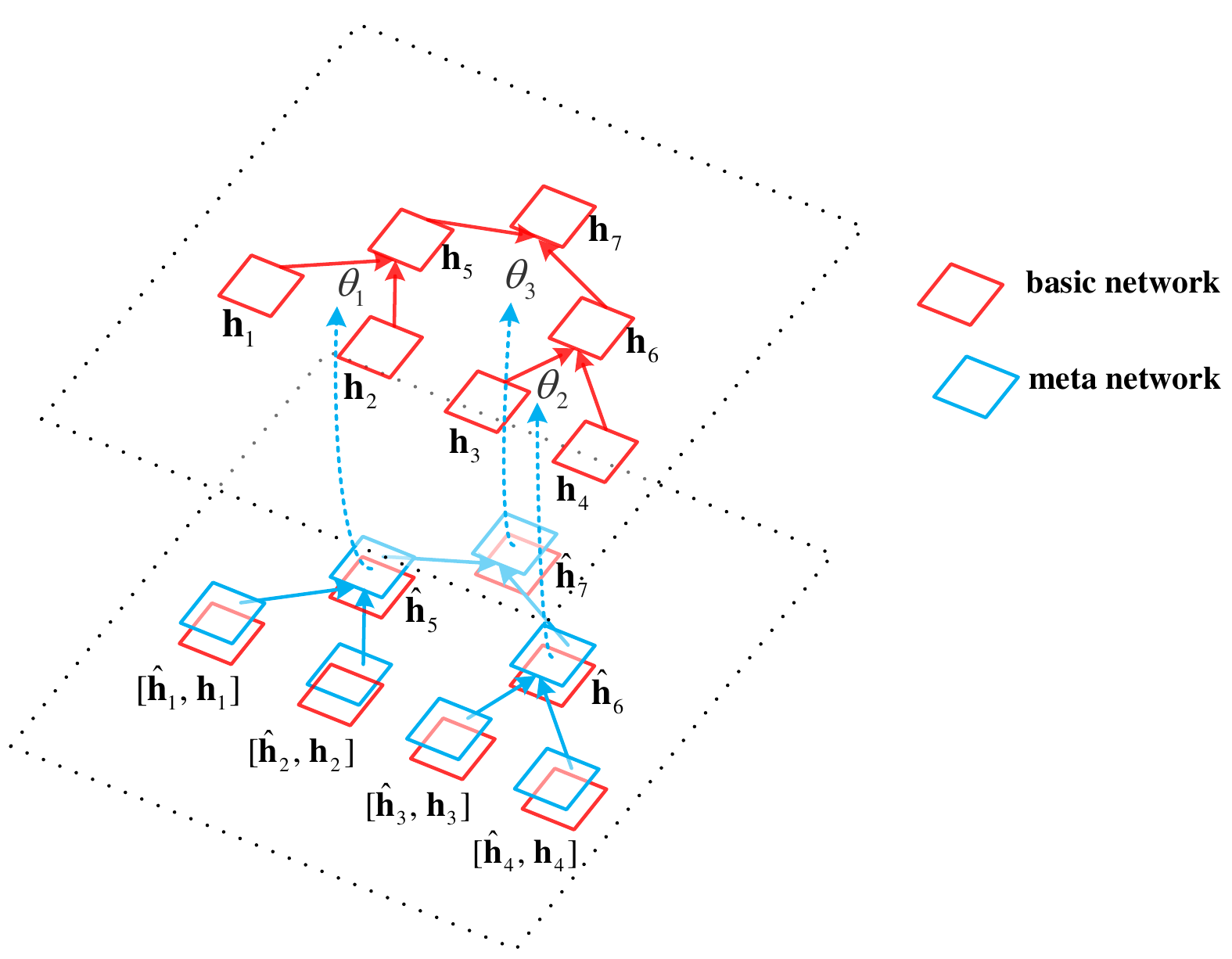}
 \caption{Dynamic Compositional Neural Network. $\theta$ denotes the context-dependent parameters generated by meta TreeNN.
 }\label{fig:case1}
\end{figure}

In the above two tree-structured NNs, the compositional function is shared across all nodes in the tree, which results in underfitting since the semantic compositions have great diversities. To address this problem, we propose two dynamic compositional neural networks over tree structure, which dynamically generate different parameters for different types of compositions. Figure \ref{fig:case1} shows an illustration of the dynamic compositional neural network, consisting of two components: (1) meta network  and (2) basic network with dynamic parameters.

Specifically, we propose two meta networks to generate the context-specific compositional functions for RecNN and TreeLSTM respectively.

\subsection{Meta Network for RecNN}
For RecNN, we replace the static parameters $\mathbf{W}$ and $\mathbf{b}$ in Eq.(\ref{eq:RecNN}) with the dynamic parameters $\mathbf{W}(\mathbf{z}_{j})$ and $\mathbf{b}(\mathbf{z}_{j})$, which are generated by a meta network.
The meta network is a smaller RecNN, and the hidden state $\hat{\h}_{j}\in \mathbb{R}^{m}$ of node $j$ in meta network is defined as
\begin{align}
    \hat{\h}_{j} &= \tanh
    \left(
	\mathbf{W}_{m}
		\mathcal{H}_{j} + \mathbf{b}_{m} \right),\\
\mathbf{z}_{j} &= \mathbf{W}_{z} \hat{\h}_{j}
\end{align}
where $\mathcal{H}_{j} = \mathbf{h}_{j}^{l}\oplus \mathbf{h}_{j}^{r}\oplus  \mathbf{\hat{h}}_{j}^{l}\oplus \mathbf{\hat{h}}_{j}^{r} \in \mathbb{R}^{2m+2d}$, $\mathbf{W}_{m} \in \mathbb{R}^{m \times (2d+2m)}$ and $\mathbf{b}_{m} \in \mathbb{R}^{m}$ are parameters of meta RecNN; $\mathbf{W}_{z} \in \mathbb{R}^{z \times m}$ is a scale matrix.

To reduce the number of the parameters, we define the dynamic parameters with a low-rank factorized representation of the weights, analogous to the Singular Value Decomposition. The dynamic parameters $\mathbf{W}(\mathbf{z}_{j})$ and $\mathbf{b}(\mathbf{z}_{j})$ of the basic RecNN are computed by:
\begin{align}
    \mathbf{W}(\mathbf{z}_{j})
	&=
	\begin{bmatrix}
		P_l \D(\z_j)Q_l \\
		P_r \D(\z_j)Q_r \\
	\end{bmatrix} \\
    \mathbf{b}(\mathbf{z}_{j})
	&=
	\begin{bmatrix}
		B_l\z_j \\
		B_r\z_j \\
	\end{bmatrix}
\end{align}
where $P \in \mathbb{R}^{d \times z}$, $Q \in \mathbb{R}^{z \times d}$, and $\D(\z_t) \in \mathbb{R}^{z \times z}$ is the diagonal matrix of $\mathbf{z}$.

Thus, our dynamic RecNN needs $(6dz + mz)$ parameters, while the vanilla RecNN has $(2d^2+d)$ parameters.
With a small $z$ and $m$, our dynamic RecNN needs less parameters than the vanilla RecNN. For example, if we set $d =100$ and $z=m=20$, our model needs $12,400$ parameters while the vanilla model needs $20,100$ parameters.

\subsection{Meta Network for TreeLSTM}
Likewise, we also use a smaller meta network to generate the static parameters $\mathbf{W}$ and $\mathbf{b}$ in Eq.(\ref{eq:treelstm1}) with the dynamic parameters $\mathbf{W}(\mathbf{z}_{j})$ and $\mathbf{b}(\mathbf{z}_{j})$.
The meta network is a smaller TreeLSTM, and the hidden state $\hat{\h}_{j}\in \mathbb{R}^{m}$ of node $j$ in meta network is defined as
\begin{align}
	\begin{bmatrix}
		\mathbf{\hat{g}}_{j} \\
		\mathbf{\hat{o}}_{j} \\
		\mathbf{\hat{i}}_{j} \\
		\mathbf{\hat{f}}_{j}^{l} \\
        \mathbf{\hat{f}}_{j}^{r}
	\end{bmatrix}
	&=
	\begin{bmatrix}
		\tanh \\
		\sigma \\
		\sigma \\
		\sigma \\
    \sigma
	\end{bmatrix}
    \left(
	\mathbf{W}_{m}
	\begin{bmatrix}
		\mathbf{x}_{j} \\
        \mathcal{H}_{j} \\
	\end{bmatrix} + \mathbf{b}_{m} \right)
    ,\label{eq:treelstm1}\\
    \mathbf{\hat{c}}_{j} &=
		\mathbf{\hat{g}}_{j} \odot \mathbf{\hat{i}}_{j}
		+ \mathbf{\hat{g}}_{j}^{l} \odot \mathbf{\hat{f}}_{j}^{l}+ \mathbf{\hat{g}}_{j}^{r} \odot \mathbf{\hat{f}}_{j}^{r} , \\
	\mathbf{\hat{h}}_{j} &= \mathbf{\hat{o}}_{j} \odot \tanh\left( \mathbf{\hat{c}}_{j} \right)\label{eq:treelstm3}, \\
    \mathbf{z}_{j} &= \mathbf{W}_{z} \hat{\h}_{j}
\end{align}
where $\mathcal{H}_{j} = \mathbf{h}_{j}^{l}\oplus \mathbf{h}_{j}^{r}\oplus  \mathbf{\hat{h}}_{j}^{l}\oplus \mathbf{\hat{h}}_{j}^{r} \in \mathbb{R}^{2m+2d}$; $\mathbf{W}_{m} \in \mathbb{R}^{5m \times (3d+2m)}$ and $\mathbf{b}_{m} \in \mathbb{R}^{m}$ are parameters of meta TreeLSTM; $\mathbf{W}_{z} \in \mathbb{R}^{z \times m}$ is a scale matrix.

The dynamic parameters $\mathbf{W}(\mathbf{z}_{j})$ and $\mathbf{b}(\mathbf{z}_{j})$ of basic TreeLSTM are computed by:
\begin{align}
    \mathbf{W}(\mathbf{z}_{j}) &= \left[\mathbf{W}^{g}, \mathbf{W}^{i}, \mathbf{W}^{f^l},\mathbf{W}^{f^r},\mathbf{W}^{o}\right] \\
    \mathbf{b}(\mathbf{z}_{j}) &= \left[\mathbf{b}^{g}, \mathbf{b}^{i}, \mathbf{b}^{f^l},\mathbf{b}^{f^r},\mathbf{b}^{o}\right],
\end{align}
where for $*\in \{c,o,i,f^l,f^r\}$,
\begin{align}
    \mathbf{W}^{*}(\mathbf{z}_{j})
	&=
	\begin{bmatrix}
		P_x^{*} \D(\z_t)Q_x^{*} \\
		P_l^{*} \D(\z_t)Q_l^{*} \\
		P_r^{*} \D(\z_t)Q_r^{*} \\
	\end{bmatrix}, \\
    \mathbf{b}^{*}(\mathbf{z}_{j})
	&=
	\begin{bmatrix}
		B_x^{*}\z_t \\
		B_l^{*}\z_t \\
		B_r^{*}\z_t \\
	\end{bmatrix},
\end{align}
where $P \in \mathbb{R}^{5d \times z}$, $Q \in \mathbb{R}^{z \times 3d}$, $B \in \mathbb{R}^{5d \times z}$, and $\D(\z_t) \in \mathbb{R}^{z \times z}$ is the diagonal matrix of $\mathbf{z}$.

%Thus, our dynamic RecNN needs $(6dz + mz)$ parameters, while the vanilla RecNN has $(15d^2+5d)$ parameters.
With a small $z$ and $m$, our dynamic TreeLSTM needs a similar amount of parameters compared to the standard TreeLSTM.

\section{Application of Dynamic Compositional Neural Networks}
In this section, we describe two specific models to show the applications of dynamic compositional neural networks for two typical tasks in NLP.

\subsection{Text Classification}
The purpose of text classification is that, given a sentence $x$, the model should predict labels $\hat{y}$ from a pre-defined label set $\mathcal{Y}$.
From the description in the previous section, we can compute the distributed representation $\h_j$ of the phrase at node $j$ of a tree:
\begin{align}
    \h_{j} &= \Basic(\x_j, \h_{j}^l, \h_{j}^r,\theta)
\end{align}
After this recursive process, the hidden state $\h_R$ at the root node is used as the sentential representation, which then followed by a softmax classifier to predict the probability distribution over classes.
\begin{align}
{\hat{\mathbf{y}}} = \softmax(\mathbf{W}_t \h_R + \bb_t)\label{eq:softmax}
\end{align}
where ${\hat{\mathbf{y}}}$ is prediction probabilities, $\W_t$ and $\bb_t$ are the parameters of the classifier.

\subsection{Text Semantic Matching}
Among many natural language processing (NLP) tasks, a common problem is modelling the relevance of a pair of texts.
In this section, we show how to effectively use the  dynamic compositional neural networks to model the semantic relationship between two sentences.

As shown in Figure \ref{fig:match}, given two sentences $x_a$ and $x_b$,
the representation of each sentence $\h_{R}$ can be computed by one basic TreeNN.
\begin{align}
    \h_{R}^{(a)} &= \Basic(\x_R, \h_{R}^l, \h_{R}^r,\theta_{a}) \\
    \h_{R}^{(b)} &= \Basic(\x_R, \h_{R}^l, \h_{R}^r,\theta_{b})
\end{align}
where $R$ denotes the root node of a tree. $\theta_{a}$ and $\theta_{b}$ are generated by a shared meta TreeNN.
Then, the representation of each sentence will be fed into a multi-layer perceptron to obtain a unified representation for the final relationship classification.

The sample-specific but shared meta TreeNN ensures that, on the one hand we can dynamically model the diversity of semantic compositionality, on the other hand we can capture the general rules across different samples.

%each sentence are encoded into a fixed representation with one basic treeNN, which are controlled by a shared meta treeNN.

\section{Experiment}
To make a comprehensive evaluation, we assess our model on five text classification tasks and a semantic matching task.%, which requires certain abilities of reasoning.

\begin{figure}[t]
\setlength{\belowcaptionskip}{-0.1cm}
\centering
 \includegraphics[width=0.43\linewidth]{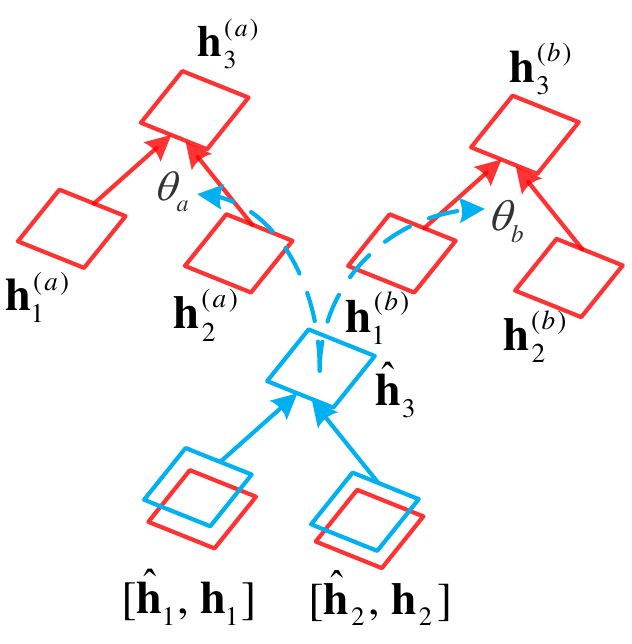}
 \caption{Illustration of DC-TreeNN Matching Network.
 }\label{fig:match}
\end{figure}

\subsection{Training and Hyperparameters}

\paragraph{Loss Function}
Given a sentence (or sentence pair) and its label, the output of a neural network is the probabilities of the different classes. The parameters of the network are trained to minimise the cross-entropy of the predicted and true label distributions.
To minimize the objective, we use stochastic gradient descent with the diagonal variant of AdaGrad \cite{duchi2011adaptive}.

\paragraph{Initialization and Hyperparameters}
The word embeddings for all of the models are initialized with GloVe vectors \cite{pennington2014glove}.
The other parameters are initialized by randomly sampling from uniform distribution in $[-0.1, 0.1]$.

The final hyper-parameters are as follows.
The initial learning rate is $0.1$. The regularization weight of the parameters is $1E{-5}$ and the others are listed as Table \ref{tab:hyper}.

For all the sentences in the datasets, we parse them with constituency parser \cite{klein2003accurate}
to obtain the trees for our models and some competitor models.

\begin{table}[t]
%\noindent $r_q$="husband" \\
\centering
\small
\begin{tabular}{cccccccc}
\toprule
 \textbf{Hyper-Param.} & {IE} & {MR} & {SST} & {SUBJ} &  {QC} & {SICK}  \\
\midrule
d   & 100 &   100 & 100 & 100 & 100 & 50(30) \\
e   & 100 &   100 & 100 & 100 & 100 & 50(50) \\
m=z & 40 &     30 & 30  & 20  & 40  & 40(20) \\
\bottomrule
\end{tabular}
\caption{Hyper-parameters for our models on all tasks. $m$, $d$ denote the size of hidden state in meta and basic TreeNN respectively.
$e$ and  $z$ represent the size of embedding vector $\x$ and controlling vector $\z$.
The settings of our two proposed models on all datasets are the same except SICK: the numbers inside and outside parentheses correspond
to DC-RecNN and DC-TreeLSTM respectively.
}
\label{tab:hyper}
\end{table}

\subsection{Competitor Methods}
\begin{itemize}
    \item RecNN \cite{socher2012semantic}: Recursive neural network with standard compositional function.
    \item RNTN \cite{socher2013recursive}: The RNTN is a recursive neural network with neural tensor layer, which can model strong interactions between two constituents.
    \item MV-RecNN \cite{socher2012semantic}: The MV-RecNN is to represent every word and longer phrase in a parse tree as both a vector and a matrix in order to model rich compositionality.
    \item TreeLSTM \cite{tai2015improved}: Recursive neural network with Long Short-Term Memory unit.
\end{itemize}

\subsection{Text Classification}

%\subsection{Datasets}
We evaluate our models on five different datasets.
The detailed statistics about the five datasets are listed in Table \ref{tab:data}. Each dataset is briefly described as follows.

\begin{table}[t]\small\centering
 \begin{tabular}{*{7}{c}}

  \toprule
  \textbf{Dataset} & Train & Dev. & Test & Class & $L_{avg}$ & $|\mathcal{V}|$\\
  \midrule
  MR    & 9596  & -     & 1066 & 2 & 22 & 21K\\
  SST   & 6920  & 872   & 1821 & 2 & 18 & 15K\\
  SUBJ  & 9000  & -     & 1000 & 2 & 21 & 21K\\
  IE    & 2221  & -     & 300 & 3 & 16 & 7.5K\\
  QC    &  5452 & -     & 500 & 6 & 10 & 9.4K\\
  \bottomrule
 \end{tabular}
 \caption{Statistics of the five mainstream datasets for text classification. $L_{avg}$
 denotes the average length of documents; $|\mathcal{V}|$ denotes the size of vocabulary.}\label{tab:data}
\end{table}

\begin{itemize*}
  \item \textbf{SST} The movie reviews with two classes (negative, positive) in the Stanford Sentiment Treebank \cite{socher2013recursive}. %\footnote{\url{http://nlp.stanford.edu/sentiment}.}
 \item \textbf{MR} The movie reviews with two classes \cite{pang2005seeing}. %\footnote{\url{https://www.cs.cornell.edu/people/pabo/movie-review-data/}.}
  \item \textbf{QC} The TREC questions dataset involves six different question types. \cite{li2002learning}. %\footnote{\url{http://cogcomp.cs.illinois.edu/Data/QA/QC/}.}
 \item \textbf{SUBJ} Subjectivity dataset where the goal is to classify each instance (snippet) as being subjective or objective. \cite{pang2004sentimental}
 \item \textbf{IE} Idiom enhanced sentiment classification. \cite{williams2015role}. Each sentence contains at least one idiom.
\end{itemize*}

\begin{table}[t]\small
\setlength{\abovecaptionskip}{0.2cm}

\center
\begin{tabular}{l*{7}{c}}
\toprule
\multicolumn{2}{c}{\textbf{Model}} &	 IE & MR &	SST & SUBJ  & QC \\	
\midrule
\multicolumn{2}{l}{NBOW}                                        & 54.6  & 77.2 & 80.5 & 91.3    & 88.2 \\
%\multicolumn{2}{l}{LSTM}                                        & 54.6  & 77.2 & 80.5 & 91.3    & 88.2 \\
\multicolumn{2}{l}{DCNN}   & -     & -    & 86.8 &	-       & 93.0 \\
\multicolumn{2}{l}{CNN-multichannel}& -     & 81.5 & \textbf{88.1} & 93.4    & 93.6 \\
\midrule
\multicolumn{2}{l}{RecNN }                 & 52.0  & 76.4 & 82.4 &	90.6    & 88.8 \\
\multicolumn{2}{l}{MV-RecNN}            & 54.8  & 76.8 & 82.9 & 90.9    & 89.2 \\
\multicolumn{2}{l}{RNTN}             & 55.7  & 75.8 & 85.4 &	92.1    & 88.9  \\
\multicolumn{2}{l}{TreeLSTM}               & 56.0  & 78.7 & 86.9 &	91.0    & 91.6 \\
\midrule
\multicolumn{2}{l}{DC-RecNN}                                  & 58.2  & 80.2 & 86.1 & 93.5    & 91.2\\
\multicolumn{2}{l}{DC-TreeLSTM}                                 & \textbf{60.2}  & \textbf{81.7} & 87.8 & \textbf{93.7}    & \textbf{93.8}\\
\bottomrule
%----------------------------------------------------------------------
\end{tabular}
\caption{
Accuracies of our models on five datasets against state-of-the-art neural models.
\textbf{DCNN}: Dynamic Convolutional Neural Network \protect \cite{kalchbrenner2014convolutional,denil2014modelling}.
\textbf{CNN-multichannel}: Convolutional Neural Network \protect \cite{kim2014convolutional}.
}\label{tab:result-movie}
\end{table}

\paragraph{Results}
As shown in Table \ref{tab:result-movie}, DC-TreeLSTM consistently outperforms RecNN, MV-RecNN, RNTN, and TreeLSTM by a
large margin while achieving comparable results to the CNN and using much fewer parameters.(The number of parameters in our models is approximately $10$K
while in CNN the number of parameters is about $400$K).
Compared with RecNN, DC-RecNN performs better, indicating the effectiveness of the dynamic compositional function.
Additionally, both DC-RecNN and DC-TreeLSTM achieve substantial improvement on IE dataset, which covers the richness of compositionality (idiomaticity).
We attribute the success on IE to its power in modeling more complicated compositionality.

\subsection{Text Semantic Matching}
We choose the dataset of Sentences Involving Compositional Knowledge (SICK), which is proposed by \newcite{marelli2014semeval} aiming at evaluation of compositional distributional semantic models. The dataset consists of 9927 sentence pairs in a 4500/500/4927 train/dev/test split, in which each sentence pairs are pre-defined into three labels: ``\texttt{entailment}'',``\texttt{contradiction}'' and ``\texttt{neutral}''.

\begin{table}[!t]\small
\small
\centering
\setlength{\belowcaptionskip}{-0.2cm}
\begin{tabular}{llll}
%\hline
\toprule
\textbf{Model} &\textbf{Hidden} & \textbf{Train} & \textbf{Test} \\
\midrule
%\hline
NBOW        & 30    & 96.6  & 73.4  \\
LSTM        & 100   & 100.0 & 71.3  \\
\midrule
RecNN       & 30    & 95.4  &  74.9 \\
MV-RecNN      & 50    & 95.9  &  75.5 \\
RNTN        & 50    & 97.8  &  76.9 \\
TreeLSTM      & 50    & 95.9  &  77.5 \\
\midrule
DC-RecNN  & 30    & 96.5    & 77.9      \\
DC-TreeLSTM & 50    & 98.5    & \textbf{80.2}   \\
\bottomrule

\end{tabular}
\caption{Evaluation results of our models on the SICK train and test sets.} \label{tab:exp-match}
\end{table}
\paragraph{Results}
Our results are summarized in Table \ref{tab:exp-match}, where the performance of NBOW, LSTM, RecNN, and RNTN are reported
by \cite{bowman-EtAl:2015:EMNLP,bowman2014recursive}.
For fair comparison, we train our models with the same setting.
We can see both DC-RecNN and DC-TreeLSTM outperform competitor models, in which DC-RecNN (DC-TreeLSTM) achieves 3\% (2.7\%) improvements
than RecNN (TreeLSTM).
We think this breakthrough is basically attributed to the dynamic compositional mechanism, which enables our models to capture various syntactic patterns (As we will discuss later)
therefore can more accurately understand sentences.

\begin{figure*}[h]
\centering
\subfloat[IE]{
  \includegraphics[width=0.19\linewidth]{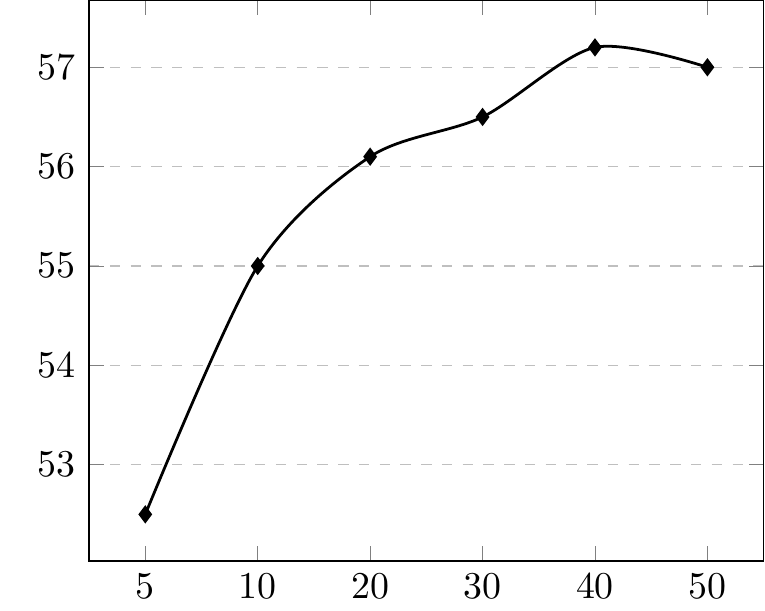}\label{fig:groups1}
  }
%  \hspace{1em}
\subfloat[MR]{
  \includegraphics[width=0.19\linewidth]{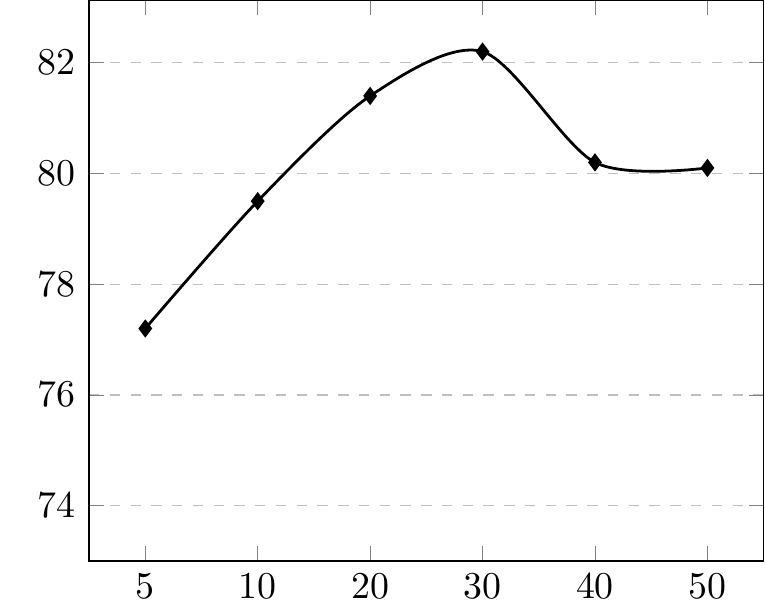}\label{fig:groups1}
  }
  %\hspace{1em}
\subfloat[SST]{
  \includegraphics[width=0.19\linewidth]{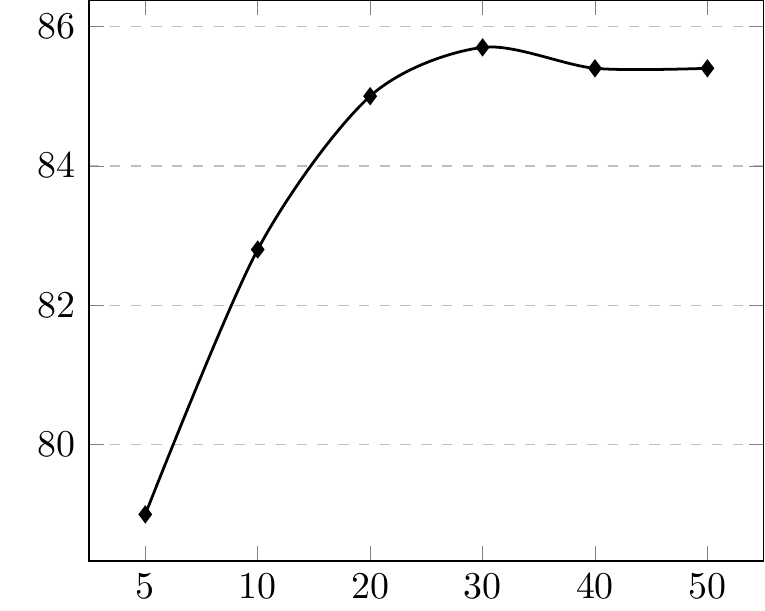}\label{fig:groups1}
  }
  %\hspace{1em}
\subfloat[SUBJ]{
  \includegraphics[width=0.19\linewidth]{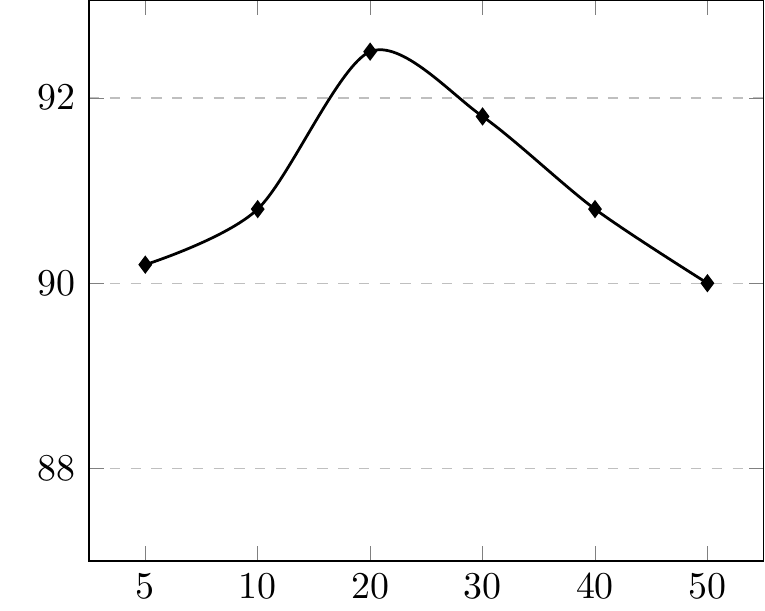}\label{fig:groups1}
  }
\subfloat[QC]{
  \includegraphics[width=0.19\linewidth]{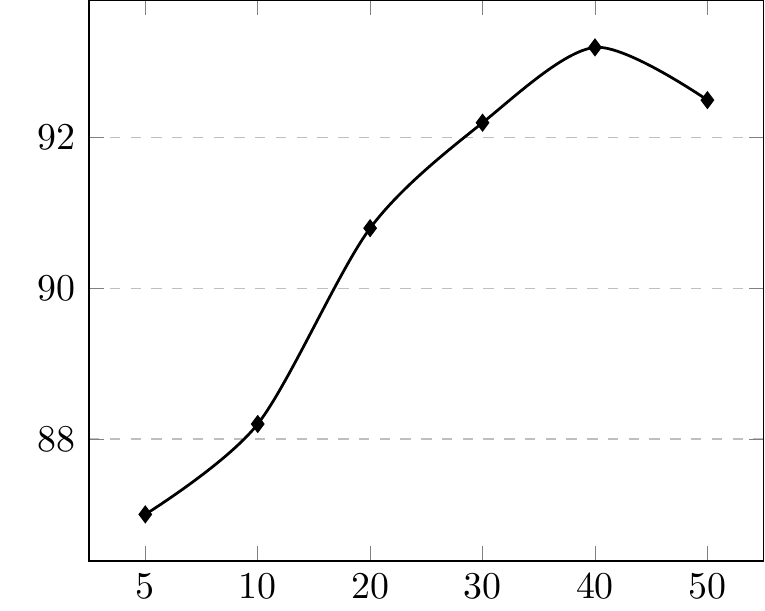}\label{fig:groups1}
  }

  \caption{Performances of DC-RecNN with the different sizes of latent vector $z$ on five development datasets: IE, MR, SST, SUBJ, and QC. Y-axis represents the accuracy(\%), and X-axis represents dimensionality of $z$.}\label{fig:dim}
\end{figure*}

\begin{table*}[!t]\small
\setlength{\belowcaptionskip}{-0.3cm}
\center
\begin{tabular}{l*{5}{l}}
\toprule
\multicolumn{2}{c}{\textbf{Type}} &	 \textbf{Neurons} &	\textbf{Examples} & \textbf{Explanations} \\
\midrule

\multirow{2}{*}{Semantic}
& Lexical                       & 17-$th$ &  fun, glad, terrific, wonderful, refreshing         & Words related to sentiment   \\
& Phrasal                       & 21-$st$ &  pick holes, see red, in stitches, split hairs      & Phrases related to sentiment     \\
\midrule
\multirow{4}{*}{Syntactic}
& Noun Phrase                    & 45-$th$ &   blond boy, pink shirt, green grass, black dog     & Containing modifiers related to color    \\
& \multirow{2}{*}{Verb Phrase}   & 27-$th$ &  waking up, take off, pulling up, driving down      & Phrases constructed by light-verb    \\
&                                & 11-$th$ &  slicing a potato, playing guitar, chopping butter  & Verb-object phrases    \\
& Prep. Phrase                   & 13-$th$ &  on a track, in rocky area, on a stage, over water  & Phrases related to places     \\
%\hline
%\multicolumn{2}{l}{{Syntactic}}
%                        &  spill the beans $\rightarrow$    beans have been spilled   \\
\bottomrule
%-----------------------------------------------------------------------------------------------------
\end{tabular}
\caption{Multiple interpretable neurons and the words/phrases captured by these neurons. The last column gives the explanations of corresponding neuron's behaviours.}\label{tab:behaviour}
\end{table*}

\subsection{Discussion and Qualitative Analysis}
In our models, the latent vector $\z$ controls the process of predicting network's parameters and its dimensionality determines the number of model's parameters. Next, we will investigate how the
controlling vector $\z$ influences the performance of our models.

\paragraph{Impact of the Dimensionality}

Figure \ref{fig:dim} shows the accuracies of DC-RecNN across the different dimensions of $[5, 10, \dots, 50]$ for the controlling vector $\z$ on five datasets.
We get the following findings:
\begin{itemize}
    \item For all five datasets, the model can achieve considerable performances even when the size of vector $\z$ is reduced to 5. Particularly, for the dataset QC, the model obtains $87.0\%$ accuracy with a pretty small meta Tree-RecNN\footnote{With the same parameters, the RecNN obtain 74\% accuracy in our implementation}, suggesting a smaller meta network can be used for generating a more powerful compositional function to effectively model sentence.
        \item
        When dealing with the dataset with more labels, larger vector size leads to a better performance. For example, the performance on IE and QC datasets reaches the maximum when the size of $z$ equals $40$, while for the other three datasets MR, SST and SUBJ, the model obtains the best performance with the value of $30$, $30$ and $20$ respectively.
\end{itemize}

\paragraph{Understanding the Neuron's Behaviours}
As described in previous sections, we know the compositional function is changed cross child nodes over a tree, which is controlled by a latent vector $z$.
To get an intuitive understanding of how the controlling vector $z$ works, we design an experiment to examine the neuron's behaviours of $\z$
on each node.
More concretely, we refer to $z_{jk}$ as the activation of the $k$-neuron at node $j$, where
$j \in \{1,\ldots,N\}$ and $k \in \{1,\ldots, z\}$.
Then we randomly sample some sentences on the development set from the datasets we used.
By visualizing the latent vector $\z_{j}$ and analyzing the maximum activation, we can find what kinds of patterns the current neuron focuses on.

\begin{figure}[!t]
\setlength{\belowcaptionskip}{-0.25cm}
  \centering
  \subfloat[The behaviour of 21-$st$ neuron for sentence ``\texttt{She had everyone in stitches}'']{
  \includegraphics[width=0.8\linewidth]{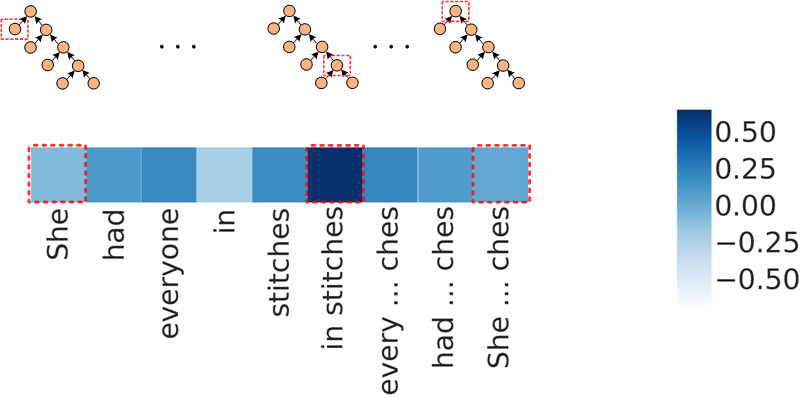}
  } \\
  \subfloat[The behaviour of 27-$th$ neuron for sentence ``\texttt{An airplane is taking off}'']{
  \includegraphics[width=0.8\linewidth]{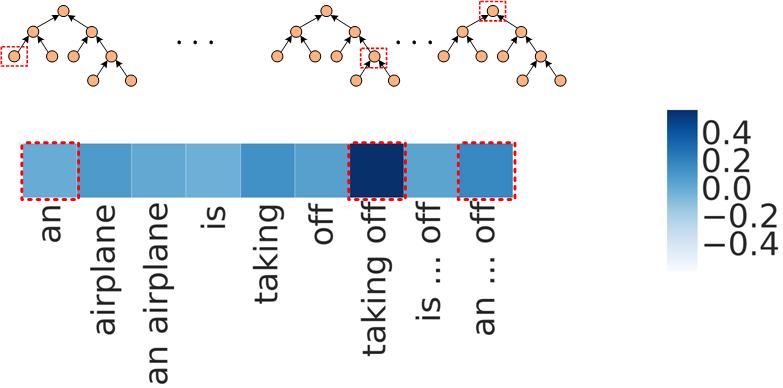}
  }
  \caption{The two heat maps describe the behaviours of neurons $\z_{21}$ and $\z_{27}$ from DC-TreeNN.}\label{fig:visual}
\end{figure}

Table \ref{tab:behaviour} illustrates multiple interpretable neurons and some representative words or phrases which can activate these neurons.
We can observe that:
\begin{itemize}
    \item For some simple tasks such as text classification,
    meta network will integrate useful semantic information into the the generation process of
    compositional function. These semantic bias before composition are task-specific.

For example, the $21$-$st$ neuron is more sensitive to emotional terms, which can be understood as a sentinel, telling the basic neural network that an informative
phrase is coming, more attention should be paid in the process of composition.
Figure \ref{fig:visual}-(a) shows a visualization. We can see in this sentence, the neuron has realized that this idiomatic collocation
``\texttt{in stitches}'' is a key pattern, which is crucial for the final sentiment prediction.

%For example,
\item For more complicated tasks such as semantic matching, a well-grounded understanding of the syntactic structure is crucial. In this context, we find that a meta network could capture some syntactic information.
For example, the $27$-$th$ neuron monitors phrases constructed by light-verb. As shown in Figure \ref{fig:visual}-(b), the verb phrase ``\texttt{taking off}'' has been
attended for forthcoming compositional operation, which is more useful for judging the semantic relation between the sentence pair
``\texttt{An airplane is taking off/A plane is landing}''.
\end{itemize}

\section{Related Work}
%A variety of recent works explore different kinds of compositional function

One thread of related work is the exploration of different kinds of compositional function over tree structures.
\newcite{socher2012semantic} proposed the recursive neural network with standard compositional function.
After that, some extensions are introduced to enhance the expressive power of compositional function, such as MV-RecNN \cite{socher2013recursive}, SU-RNN \cite{socher2013parsing}, RNTN \cite{socher2013recursive},  while these models suffer from the problem of hard-coded compositional operations and overfitting.

Another thread of work is the idea of using one network to direct the learning of another network \cite{de2016dynamic}.
\newcite{naik1992meta} introduce a meta neural network to provide another network with a step size and a direction vector, which is helpful for parameter optimization.
\newcite{de2016dynamic} propose the dynamic filter network to implicitly learn a variety of filtering operations.
\newcite{bertinetto2016learning} introduce a learnet for one-shot learning, which can predict the parameters of a second network given a single exemplar.
\newcite{ha2016hypernetworks} propose the model hypernetwork, which uses a small network to generate the weights for a larger network.

Different from these models, we employ the idea of parameter generation to address the limitation of weight-sharing or partially sharing paradigm of tree-based compositional models.

\section{Conclusion}

In this work, we introduce a meta neural network, which can generate a compositional network to dynamically compose constituents over tree structure. The parameters of compositional function vary from position to position and from sample to sample, allowing for more sophisticated operations on the input.

To evaluate our models, we choose two typical NLP tasks involving six datasets. The qualitative and quantitative experiment results demonstrate the effectiveness of our models.

\section*{Acknowledgments}
We would like to thank the anonymous reviewers for their valuable comments and thank Kaiyu Qian, Jiachen Xu, Jifan Chen for useful discussions.
This work was partially funded by National Natural Science Foundation of China (No. 61532011 and 61672162), Shanghai Municipal Science and Technology Commission (No. 16JC1420401).

\bibliographystyle{named}
\bibliography{nlp1,ours1,nlp,ours2}

\end{document}